%% file: sample-acmtog-SIGGRAPH-submission.tex
\documentclass[acmtog,nonacm]{acmart}
\authorsaddresses{}
\makeatletter
\let\@ACM@checkaffil\relax
\makeatother
\usepackage{xcolor}
\usepackage{pifont}
\usepackage{makecell}
\usepackage{multirow}
\definecolor{revisionblue}{RGB}{0.00,0.00,0.00}
\newif\ifshowrevisions
\showrevisionstrue
\DeclareRobustCommand{\rev}[1]{\ifshowrevisions{\color{revisionblue}#1}\else#1\fi}
\newcommand{\revcolor}{\ifshowrevisions\color{revisionblue}\fi}
\usepackage[ruled]{algorithm2e} 

\SetAlFnt{\small}
\SetAlCapFnt{\small}
\SetAlCapNameFnt{\small}
\SetAlCapHSkip{0pt}

\begin{document}
\title{\rev{Test-Time Scaling for Video Diffusion Models via Diagnosis-Guided Candidate Recycling}}

\author{Hangzhou He}
\authornote{Work done during an internship at Alibaba Group.}
\affiliation{\institution{Peking University} \country{China}}
\affiliation{\institution{Alibaba Group} \country{China}}

\author{Lunhao Duan}
\affiliation{\institution{Alibaba Group} \country{China}}

\author{Shanshan Zhao}
\authornote{Corresponding author. Email: sshan.zhao00@gmail.com, yanye.lu@pku.edu.cn.}
\affiliation{\institution{Alibaba Group} \country{China}}
\email{sshan.zhao00@gmail.com}

\author{Kaiwen Li}
\affiliation{\institution{Peking University} \country{China}}

\author{Qing-Guo Chen}
\affiliation{\institution{Alibaba Group} \country{China}}

\author{Weihua Luo}
\affiliation{\institution{Alibaba Group} \country{China}}

\author{Yanye Lu}
\authornotemark[2]
\affiliation{\institution{Peking University} \country{China}}
\email{yanye.lu@pku.edu.cn}

\renewcommand\shortauthors{He et al.}
\begin{abstract}
Recent video diffusion models have achieved remarkable generation quality, but high-fidelity results still largely depend on closed-source systems or costly large-scale infrastructure. Test-time scaling (TTS) offers a training-free way to improve lightweight generators by spending additional inference compute, yet existing methods mostly remain within a noise-search paradigm: they sample, select, or perturb denoising trajectories and discard low-scoring candidates after expensive generation. This generate-and-discard process wastes not only computation but also the partial motion, layout, or appearance structure already encoded in recoverable samples. We present \textbf{GEARS} \rev{(\textbf{G}uided \textbf{E}diting for \textbf{A}daptive \textbf{R}ecycling \textbf{S}earch)}, a training-free framework that \rev{introduces {diagnosis-guided candidate recycling} into video TTS by turning} such candidates into editable priors through a generation-evaluation-editing loop. GEARS consists of two collaborative components. \rev{The \textbf{Stage-Aware Scheduler} determines what to repair, when to repair it, and which candidates should be preserved, recycled, or discarded. The \textbf{Candidate Recycler} diagnoses recoverable failures from keyframes and multi-dimensional reward feedback, derives candidate-specific repair prompts, and repairs the corresponding candidates through manifold-aware latent SDEdit.} The repaired candidates are recycled into the search pool, creating refinement paths beyond standard noise perturbation while preserving useful structure. Under matched NFE budgets, GEARS consistently outperforms existing video TTS methods on VBench, bringing a 1.3B model to a total score comparable to a 14B counterpart, and ablations verify the necessity of adaptive scheduling, diagnosis-conditioned editing, and manifold-aware re-denoising. Code is available at \rev{https://github.com/ATH-MaaS/Guided\_Editing\_for\_Adaptive\_Recycling\_Search}. 
\end{abstract}


\begin{teaserfigure}
    \centering
    \includegraphics[width=\textwidth]{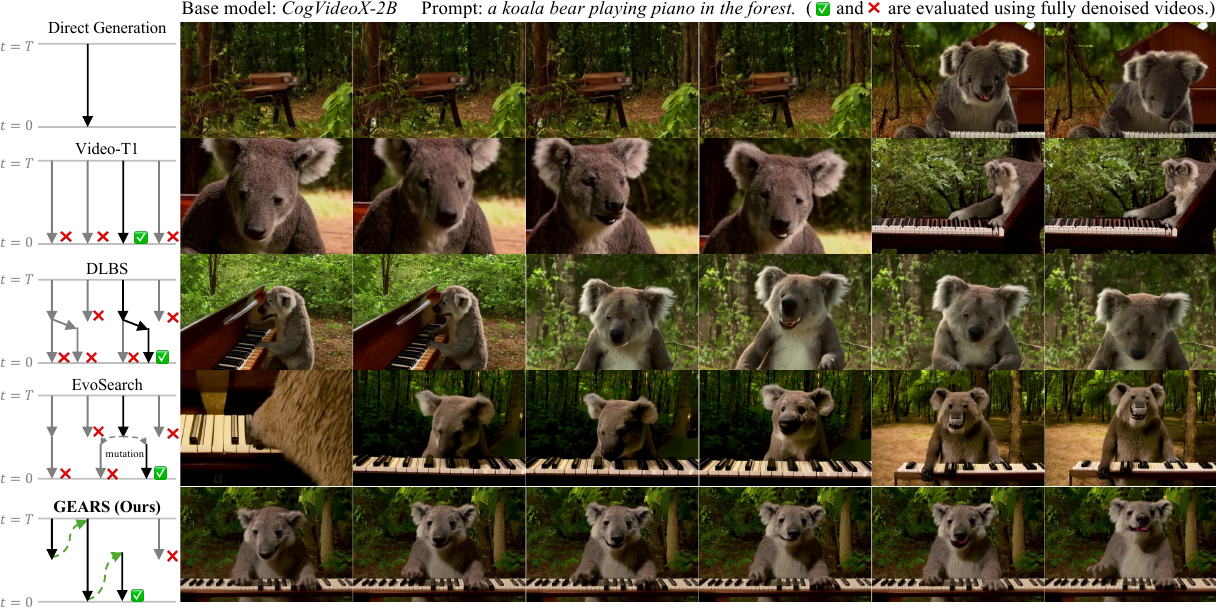}
    \caption{\textbf{Left:} Test-time scaling paradigms along the denoising axis ($t{=}1 \!\to\! t{=}0$). Video-T1~\cite{liu2025video} samples independent trajectories and selects the best video at the end; DLBS~\cite{oshima2025inference} seeks a better diffusion path by keeping K elites for each step as beam search; EvoSearch~\cite{he2025scaling} adds mid-denoising mutation around elites but still clusters candidates locally; \textbf{GEARS} further recycles low-scoring candidates via diagnostic latent editing (green dashed arrows), rerouting them back into denoising at two scheduled stages to reach regions noise-space search alone cannot. \textbf{Right:} Qualitative comparison. Baselines suffer from subject-background inconsistency or stagnant motion, while GEARS produces a coherent koala playing piano throughout.}
    \label{fig:teaser}
\end{teaserfigure}

\maketitle

\input{samplebody-journals}

\end{document}

%% file: samplebody-journals.tex
\section{Introduction}\label{sec:introduction}
Recent advancements in video generation heavily rely on scaling model capacity and extensive training datasets \cite{liu2024sora,klingAI,wan2025wan,google2025veo3,seedance2026seedance}. While large closed-source models achieve impressive cinematic quality, their prohibitive computational demands drive a parallel effort to develop lightweight models suitable for consumer hardware \cite{hong2022cogvideo,yang2024cogvideox,zheng2025open,wan2025wan,wu2025hunyuanvideo}. However, a substantial quality gap persists between open-source models and larger counterparts. 

Test-time scaling (TTS) provides an opportunity to bridge this gap by allocating additional computation during inference without altering model weights. Inspired by search strategies in large language models \cite{snell2024scaling,chen2024expanding}, recent TTS methods for diffusion models search over prompts, initial noises, or denoising trajectories with reward-based verifiers \cite{ma2025inference}. Prompt-space methods refine text conditions before generation \cite{long2025vista,gao2025rapo,song2026vqqa}, whereas generation-space methods explore denoising trajectories through discrete sampling \cite{liu2025video}, evolutionary optimization \cite{he2025scaling}, beam search \cite{oshima2025inference}, or semantic budget allocation \cite{wu2026imagerysearch}. However, these methods largely retain a generate-evaluate-discard paradigm, where lower-ranked candidates are discarded after expensive denoising and evaluation.

\begin{figure}[t]
    \centering
    \includegraphics[width=\columnwidth]{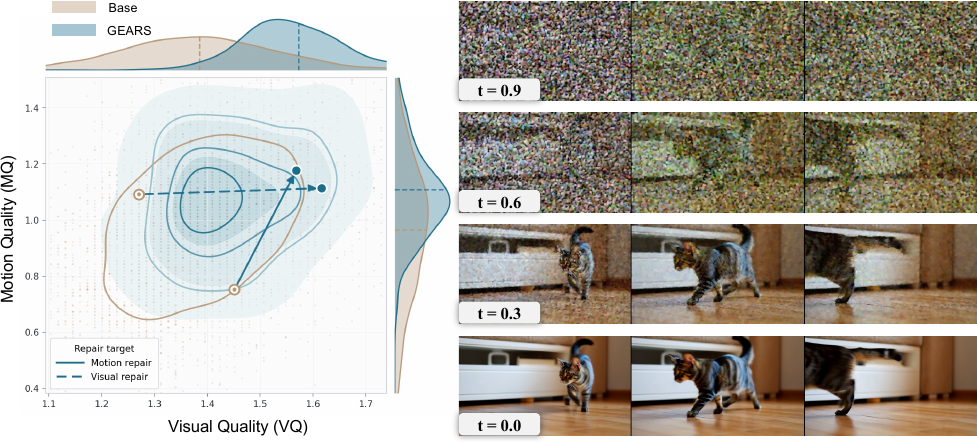}
    \caption{\textbf{Left}: \rev{Joint and marginal distributions of Base and GEARS candidates in the VQ--MQ space. Contours visualize the joint density, while the top and right curves show the VQ and MQ marginal profiles, respectively. GEARS redirects recoverable candidates toward higher joint quality.} \textbf{Right}: High-noise stages establish coarse motion and layout, whereas low-noise stages refine appearance and texture.}
    \label{fig:denoising_process}
\end{figure}

However, video generation is inherently open-ended and subjective: a plausible video may still receive a low reward because it underperforms along a particular aspect such as visual fidelity, motion dynamics, or text alignment \cite{liu2025improving}. As shown in Fig.~\ref{fig:denoising_process} (left), a low overall score does not necessarily indicate a globally defective sample: many low-scoring candidates remain plausible in some dimensions while failing predominantly along one recoverable aspect \cite{luo2026beyond}. Discarding such candidates therefore wastes both the useful structure they encode and the computation already spent on their generation.

Furthermore, existing search methods obtain diversity mainly from independent initial noises or local stochastic perturbations around promising trajectories (Fig.~\ref{fig:teaser} left). Such exploration remains confined to the forward noise-search process, causing candidates to cluster around similar denoising basins and leading to rapidly diminishing returns as more inference compute is spent. Although diffusion models naturally support training-free editing through noise-and-redenoise procedures \cite{meng2021sdedit,lugmayr2022repaint}, this capability remains largely underexplored as an integral component of TTS for text-to-video generation.

\rev{In this paper}, we introduce \textbf{GEARS} \rev{(\textbf{G}uided \textbf{E}diting for \textbf{A}daptive \textbf{R}ecycling \textbf{S}earch)}, a training-free framework that reframes low-scoring but structurally plausible candidates as editable priors rather than failed samples. Instead of continuing the generate-evaluate-discard loop, GEARS diagnoses the dimension-specific failures of these candidates, repairs them by editing their re-noised latent states at the denoising stage where the target attribute is most controllable, and recycles the corrected candidates back into the search pool. \rev{We refer to this diagnose-repair-recycle mechanism as {diagnosis-guided candidate recycling}, which realizes a generation-to-editing transition by converting wasted computation into structural diversity and shifting test-time compute from passive noise selection to active refinement.} \rev{GEARS realizes this transition through two collaborative components: the \textbf{Stage-Aware Scheduler} (Scheduler) and the \textbf{Candidate Recycler} (Recycler).}

\rev{The \textbf{Scheduler} determines what to repair, when to repair it, and which candidates should be preserved, recycled, or discarded. It uses the denoising hierarchy to assign motion and coarse-structure repair to high-noise stages and visual refinement to low-noise stages. At each stage, it preserves high-scoring candidates as elites, discards severely misaligned samples, and sends semantically valid but dimensionally deficient candidates to the Recycler.}

\rev{The \textbf{Recycler} diagnoses and repairs the recoverable candidates selected by the Scheduler and returns them to the search pool. Given multi-dimensional reward and visual anchors from the video, an MLLM diagnoses the primary failure mode and produces a candidate-specific repair instruction. The Recycler then re-noises the candidate latent to the scheduled denoising stage and re-denoises it under the repair condition, correcting the diagnosed deficiency while preserving useful motion, layout, or appearance. We further use a manifold-aware SDE (MA-SDE) \cite{zheng2026manifold} during re-denoising to reduce off-manifold artifacts and temporal jitter.}

Together, this structurally aware generation-evaluation-editing loop moves beyond the local neighborhoods explored by standard noise perturbation and opens new refinement paths through edited structural priors. Quantitative and qualitative evaluations show that GEARS consistently outperforms existing noise-space TTS methods under matched compute budgets, providing a new generation-to-editing paradigm that shifts the focus of video test-time scaling from passive noise selection to active, structured refinement.

\section{Related Work}\label{sec:relatedwork}
\subsection{Video Generation and Editing}\label{sec:relatedwork_video_models}
Diffusion- and flow-matching-based video generation \cite{dhariwal2021diffusion,lipman2023flow} has advanced rapidly, driven by progress in model architectures \cite{liu2022flow,peebles2023scalable} and training-time scaling \cite{yu2025context,he2025cut2next,gu2025diffusion}. Large-scale systems such as Kling \cite{klingAI}, Veo \cite{google2025veo3}, and Seedance \cite{seedance2026seedance} have demonstrated cinematic-quality generation, while open-source models \cite{yang2024cogvideox,zheng2025open,wan2025wan,kong2024hunyuanvideo,wu2025hunyuanvideo} have made text-to-video generation increasingly accessible on consumer hardware. Despite this progress, small models still lag behind their large-scale counterparts, motivating inference-time strategies that improve generation quality without additional training or model-size scaling. 

A complementary line of work studies training-free video editing. Diffusion and flow-matching models naturally support editing through their forward-reverse generative structure. SDEdit \cite{meng2021sdedit} shows that adding noise to an existing sample and re-denoising it under new conditioning can produce meaningful edits without fine-tuning. Recent video editing methods further exploit this property through noise inversion, clean-latent manipulation, localized noise splicing, or context-aware refinement \cite{lee2024videorepair,bai2025uniedit,samuel2025omnimattezero,chen2026contextflow}. These works demonstrate the built-in capacity for refinement and correction.

However, existing editing methods are typically formulated as standalone editing or post-processing procedures applied to a given output. Their correction ability remains largely disconnected from test-time scaling, where additional compute is instead spent on sampling, scoring, and selecting candidates. GEARS bridges this gap by embedding MLLM-guided latent-space editing directly into the inference-time search loop. Rather than editing only a final output, GEARS diagnoses low-scoring but structurally plausible candidates during search, repairs them through latent re-denoising, and recycles the corrected candidates back into the candidate pool.

\subsection{Test-Time Scaling for Video Generation}\label{sec:relatedwork_tts}
Test-time scaling (TTS) improves model outputs by allocating additional computation during inference. Originally studied extensively in LLMs \cite{snell2024scaling}, TTS commonly takes the form of parallel sampling, such as Best-of-N, or long sequential reasoning chains \cite{wei2022chain}. TTS has also been extended to image generation with diffusion models \cite{song2021scorebased,ma2025inference}. However, simply increasing the number of denoising steps is often ineffective and may even degrade visual quality due to accumulated errors \cite{ma2025inference}. Instead, diffusion TTS methods typically improve outputs by searching over prompts \cite{qu2026scale,wang2026imagentunifiedmultimodalagent}, initial noises \cite{wang2025silent,guo2024initno}, or denoising trajectories with reward-based verifiers \cite{ren2025scale}.

Video TTS largely follows the design dimensions of \emph{verifiers} and \emph{search algorithms}, but its high-dimensional and temporal nature introduces additional challenges beyond image generation. Each candidate is substantially more expensive to generate and evaluate, while its quality depends on multiple coupled factors such as visual fidelity, motion dynamics, temporal consistency, and text alignment \cite{liu2025improving,wu2025rewarddance}. This makes the generate-and-discard strategy inefficient for video generation, since low-ranked candidates may still contain useful structure or motion patterns that have already consumed significant inference compute.

Along the verifier axis, Video-T1 \cite{liu2025video} shows that ensembling complementary metrics is more effective than relying on a single reward, while \citet{baraldi2025verifier} demonstrates that stronger reward models can yield disproportionate gains under inference-time scaling. Unlike verifiable domains such as mathematics or programming \cite{chen2026think,li2025s}, video generation is open-ended and subjective, making reward construction substantially more challenging \cite{he2024videoscore,he2025videoscore2,xu2026visionreward,liu2025improving}. Recent work therefore studies more scalable and efficient reward mechanisms, including reward scaling for visual generation \cite{wu2025rewarddance} and latent reward models that score noise directly without full VAE decoding \cite{zhao2026latsearch}.

Along the search algorithm axis, prompt-space methods such as VISTA \cite{long2025vista}, VQQA \cite{song2026vqqa}, and RAPO$++$ \cite{gao2025rapo} iteratively rewrite prompts to improve generation quality, which is suitable for closed-source systems where only prompt-level access is available. For noise-space search, Video-T1 \cite{liu2025video} applies Best-of-N sampling to validate test-time scaling laws for video generation and proposes a Tree-of-Frames strategy for more efficient scaling in autoregressive video generation. Diffusion Latent Beam Search (DLBS) \cite{oshima2025inference} maintains multiple candidate trajectories with lookahead estimation, showing that reliable candidate assessment may require multi-step or full denoising. EvoSearch \cite{he2025scaling} formulates noise-space exploration as evolutionary optimization with selection, crossover, and mutation. ImagerySearch \cite{wu2026imagerysearch} adaptively allocates search budgets according to prompt complexity and semantic dependency.

Despite their effectiveness, most video TTS methods still follow a generation-and-discard paradigm, where unselected candidates are discarded after costly denoising and evaluation. GEARS departs from this paradigm by recycling recoverable low-scoring candidates through diagnosis-conditioned latent editing, turning already-spent inference compute into structured diversity for subsequent search.

\begin{figure*}[t]
    \centering
    \includegraphics[width=\textwidth]{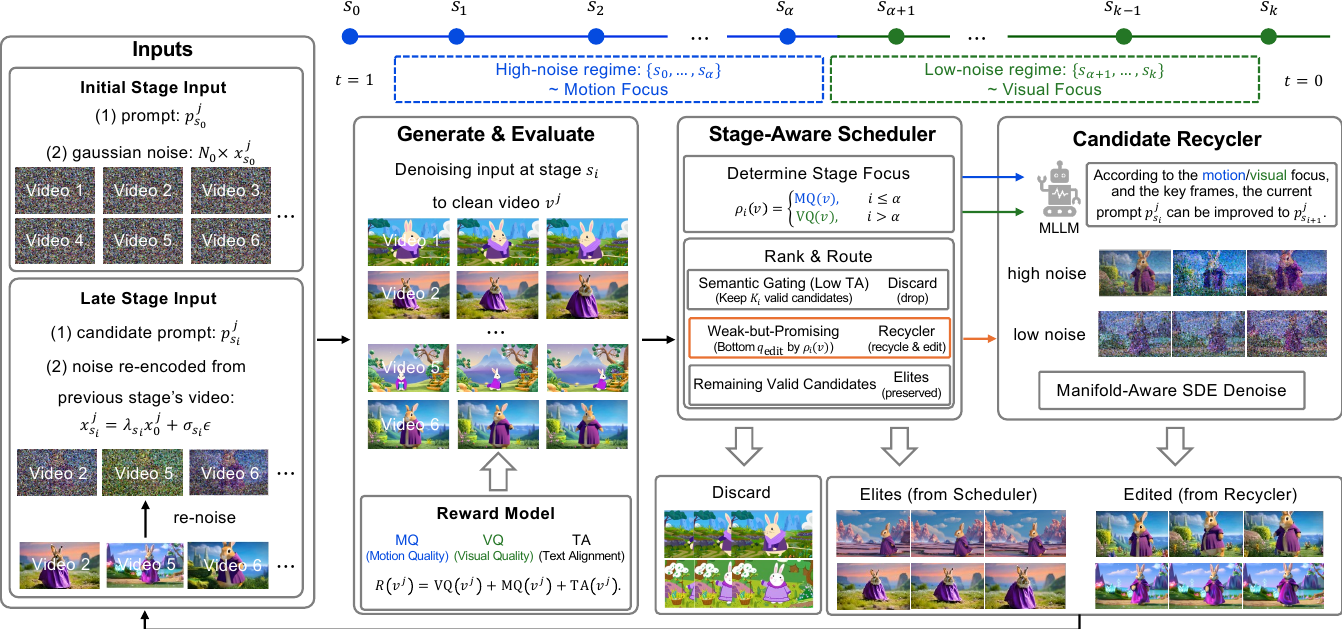}
   \caption{\textbf{Overview of GEARS.} Along the denoising trajectory from pure noise to clean latent, GEARS inserts multiple generation-to-editing checkpoints $\{s_0,s_1,\ldots,s_k\}$. \rev{At each checkpoint, candidates are first fully denoised, decoded, and evaluated as clean videos; noisy latents are never scored. The Scheduler determines the current repair focus and routes the evaluated candidates into preserved elites, recoverable candidates for the Recycler, and discarded samples. The Recycler diagnoses and repairs the recoverable candidates through prompt-guided latent editing, after which they are returned to the search pool via re-noising. High-noise checkpoints prioritize motion correction, while low-noise checkpoints focus on visual refinement.}}\label{fig:method_overview}
\end{figure*}

\section{Methodology}\label{sec:methodology}
\subsection{Overview of GEARS}\label{sec:method_overview}
Consider a pretrained latent text-to-video generator $\mathcal{G}$ that operates in the latent space. Given a text prompt $p$ and an intermediate latent $x_t$ at noise level $t \in [0,1]$ (where $t=1$ denotes pure noise and $t=0$ denotes a clean latent), the generator defines a conditional reverse denoising process:
\begin{equation}
    x_{t'} = \mathcal{G}_{t \rightarrow t'}(x_t, p),
    \qquad 0 \leq t' < t \leq 1.
\end{equation}
The clean latent $x_0$ is decoded into a video $v$ by the decoder ${\rm Dec}(\cdot)$ of VAE. GEARS also uses the standard forward perturbation operation associated with the same generator schedule. Given a clean latent $x_0$, we can perturb it back to a target noise level $t$ as
\begin{equation}
    x_t = \lambda_t x_0 + \sigma_t \epsilon,
    \qquad \epsilon\sim\mathcal{N}(0,I), \label{eq:renoise}
\end{equation}
where $\lambda_t$ and $\sigma_t$ are the signal and noise coefficients determined by the diffusion or flow-matching sampler schedule.

As shown in Fig.~\ref{fig:method_overview}, GEARS selects a sequence of discrete noise levels as generation-to-editing checkpoints along the denoising trajectory:
\begin{equation}
    1=s_0 > s_1 > \cdots > s_k > 0.
\end{equation}
Each $s_i$ is a specific noise level which \rev{defines an outer-loop recycling stage. Every candidate is first rolled out from $s_i$ to $0$, decoded, and scored as a clean video, while $s_i$ controls the editing strength and target attributes.} \rev{This process is implemented by two components: the Scheduler determines the repair focus and routes candidates at each checkpoint, while the Recycler diagnoses and repairs the selected recoverable candidates.}
Here, we first provide a brief overview of this generation-evaluation-editing loop.

At checkpoint $s_i$, GEARS maintains a candidate pool $\mathcal{P}_i$ consisting of latent-prompt pairs $(x_{s_i}^j,p^j_{s_i})$. The first checkpoint starts from Gaussian noise:
\begin{equation}
    \mathcal{P}_0=\{(x_{s_0}^j,p^j_{s_0})\}_{j=1}^{N_0}, \qquad x_{s_0}^j\sim\mathcal{N}(0,I),
\end{equation}
where $N_0$ is the initial sample size. While all initial prompts $p^n_{s_0}$ equal the original prompt $p$, subsequent prompts $p^j_{s_i}$ become candidate-specific, as \rev{the Recycler} may modify them to guide targeted repairs. For each candidate in $\mathcal{P}_i$, we run the generator from $s_i$ to $0$ and decode the resulting clean latent:
\begin{equation}
    x_0^j=\mathcal{G}_{s_i\rightarrow 0}(x_{s_i}^j,p^j_{s_i}), \qquad v^j=\mathrm{Dec}(x_0^j).
\end{equation}

Each generated video $v^j$ is then evaluated by a multi-dimensional reward model along visual quality (VQ), motion quality (MQ), and text alignment (TA). \rev{Here we use VideoAlign \cite{liu2025improving} with its default unweighted sum} as the global ranking score, \rev{without tuning it on VBench or human preferences}:
\begin{equation}
    R(v^j) = \mathrm{VQ}(v^j) + \mathrm{MQ}(v^j) + \mathrm{TA}(v^j).
\end{equation}

After evaluation, \rev{the Scheduler} routes the candidates into three groups, as illustrated in Fig.~\ref{fig:method_overview}. Candidates with poor text alignment are discarded by a semantic gate; among the remaining candidates, those with low scores under the current stage focus are sent to the generation-to-editing branch, while the others are preserved as elites, \rev{denoted by $\mathcal{E}_i$. The Recycler diagnoses and repairs the selected candidates through prompt-guided re-denoising, producing next-checkpoint prompts $p_{s_{i+1}}^j$ for the repaired candidates.} The preserved elites and repaired candidates are then merged into a survivor set $\mathcal{B}_i$, which stores each surviving clean latent together with the prompt to be used at the next checkpoint. For preserved elites, the prompt is carried over unchanged, i.e., $p_{s_{i+1}}^j = p_{s_i}^j$.

\rev{For $i<k$,} the non-elite survivors in $\mathcal{B}_i \setminus \mathcal{E}_i$ are perturbed to the next checkpoint $s_{i+1}$ by Eq.~(\ref{eq:renoise}) to form the next candidate pool:
\begin{equation}
    \mathcal{P}_{i+1} = \{(x_{s_{i+1}}^j,{p}^j_{s_{i+1}})\mid ({x}_0^j,{p}^j_{s_{i+1}})\in\mathcal{B}_i \setminus \mathcal{E}_i\}.
\end{equation}
Preserved elites retain their clean latents, inherited prompts, and reward scores, and are directly merged with the newly evaluated candidates at checkpoint $s_{i+1}$ for \rev{Scheduler routing}. GEARS repeats this loop at each subsequent checkpoint until the final output is selected by the ranking score $R$ at the last checkpoint.

\subsection{\rev{Stage-Aware Scheduler}}
\label{sec:stageadapter}
\rev{The Scheduler determines what to repair, when to repair it, and which candidates to discard, preserve, or recycle at each generation-to-editing checkpoint.} Its design follows the hierarchical nature of the denoising process: earlier checkpoints with higher noise levels have stronger influence on global layout and motion, whereas later checkpoints with lower noise levels mainly refine appearance, texture, and local visual fidelity (Fig.~\ref{fig:method_overview} right). To operationalize this hierarchy, we introduce an empirically chosen transition index $\alpha$ over the checkpoint sequence $\{s_1,\ldots,s_k\}$, which partitions the checkpoints into a high-noise regime $\{s_0,\ldots,s_\alpha\}$ and a low-noise regime $\{s_{\alpha+1},\ldots,s_k\}$. \rev{The Scheduler} then assigns a stage-dependent editing focus:
\begin{equation}
    \rho_i(v)=
    \begin{cases}
        \mathrm{MQ}(v), & i\leq \alpha,\\
        \mathrm{VQ}(v), & i>\alpha.
    \end{cases}
\end{equation}
Here, $\rho_i(v)$ indicates the attribute to prioritize when selecting recoverable candidates for editing at checkpoint $s_i$: motion quality in the high-noise regime and visual quality in the low-noise regime.

\rev{The Scheduler routes candidates through a rank-based procedure rather than a fixed-score threshold.} At checkpoint $s_i$, let $K_i$ denote the number of candidates to be kept for the next checkpoint. Given the evaluated candidate set, \rev{the Scheduler} first discards candidates with the lowest text-alignment scores until $K_i$ candidates remain. \rev{Let $\mathcal{C}_i$ denote this set of remaining valid candidates.} This semantic filtering removes samples whose content is missing, incorrect, or too far from the prompt, for which structure-preserving editing is usually unreliable \cite{meng2021sdedit,lee2024videorepair}.

\rev{The Scheduler} then ranks $\mathcal{C}_i$ by the current stage focus $\rho_i$, which serves as the single-dimensional criterion for deciding whether a candidate should be repaired at checkpoint $s_i$. We introduce an editing ratio $q_{\mathrm{edit}}\in[0,1]$: the bottom $q_{\mathrm{edit}}$ fraction under $\rho_i$ is selected as weak-but-promising candidates
\begin{equation}
    \mathcal{W}_i = \operatorname{Bottom}_{q_{\mathrm{edit}}} \bigl(\mathcal{C}_i; \rho_i\bigr),
\end{equation}
and sent to \rev{the Recycler for diagnosis and targeted repair}. The remaining candidates are directly preserved as elites,
\begin{equation}
    \mathcal{E}_i = \mathcal{C}_i \setminus \mathcal{W}_i .
\end{equation}
The edited candidates and preserved elites are merged to form the survivor set for the next checkpoint.

\begin{table*}[tb]
    \caption{\textbf{Quantitative Comparison Results.} (Results of ImagerySearch are from \cite{wu2026imagerysearch})}
    \label{tab:vbench_results}
    \begin{minipage}{\textwidth}
    \centering
    \begin{tabular}{lccccccccc}
        \toprule
        & \multicolumn{7}{c}{\textbf{VBench}} & \multicolumn{2}{c}{\textbf{VideoAlign}} \\
        \cmidrule(lr){2-8} \cmidrule(lr){9-10}
        \textbf{Method} & \makecell{Total\\Score} 
        & \makecell{Quality\\Score} 
        & \makecell{Semantic\\Score} 
        & \makecell{Background\\Consistency} 
        & \makecell{Imaging\\Quality} 
        & \makecell{Motion\\Smoothness}
        & \makecell{Subject\\Consistency}
        & \makecell{Visual\\Quality} 
        & \makecell{Motion\\Quality}\\
        \midrule
        Wan2.1-T2V-14B & 0.8369 & 0.8559 & 0.7611 & 0.9809 & 0.6943 & 0.9830 & 0.9752 & 1.3807 & 0.9499\\
        HunyuanVideo 13B & 0.8324 & 0.8509 & 0.7582 & 0.9776 & 0.6756 & 0.9899 & 0.9737 & 1.4240 & 1.0920\\
        \midrule
        Wan2.1-T2V-1.3B & 0.8192 & 0.8471 & 0.7076 & 0.9655 & 0.6844 & 0.9674 & 0.9687 & 1.3737 & 0.9552\\
        $+$ Video-T1 & 0.8294 & 0.8490 & 0.7508 & 0.9656 & 0.6832 & 0.9644 & 0.9684 & 1.4464 & 1.1222\\
        $+$ DLBS & 0.8224 & 0.8483 & 0.7188 & 0.9485 & 0.7039 & 0.9619 & 0.9627 & 1.4405 & 1.0834\\
        $+$ EvoSearch & 0.8190 & 0.8478 & 0.7038 & 0.9483 & 0.6951 & 0.9585 & 0.9562 & 1.4609 & 1.1009\\
        $+$ ImagerySearch & - & - & - & 0.9600 & 0.6920 & 0.9800 & 0.9590 & - & -\\
        $+$ GEARS (\textbf{Ours}) & 0.8375 & 0.8553 & 0.7661 & 0.9725 & 0.6960 & 0.9727 & 0.9815 & 1.4733 & 1.1064\\
        \bottomrule
    \end{tabular}
    \end{minipage}
    \footnotesize
\end{table*}

\begin{figure}[t]
    \centering
    \includegraphics[width=\columnwidth]{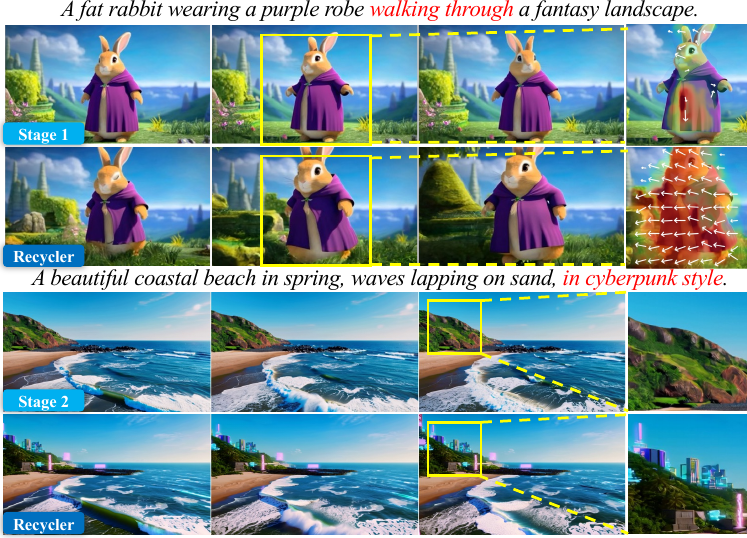}
    \caption{\textbf{Stage-aware candidate repair by \rev{the Scheduler} and \rev{Recycler}.} \textbf{Top:} At \rev{Stage} 1 (high noise), \rev{the candidate under-realizes the requested walking motion.} \rev{The Recycler restores coherent subject motion while retaining the rabbit and the fantasy scene. The rightmost insets summarize background-compensated optical flow across the three frames under a shared scale; warmer colors indicate larger residual motion, and arrows show its direction.} \textbf{Bottom:} At \rev{Stage} 2 (low noise), the candidate preserves the beach layout and wave dynamics but misses the requested cyberpunk appearance. \rev{The Recycler} introduces neon structures while maintaining the coastal composition, as highlighted by the enlarged regions. Together, the two cases illustrate how \rev{the Scheduler} assigns motion and appearance deficiencies to their appropriate repair stages.}
    \label{fig:effect_of_recycler}
\end{figure}

\subsection{\rev{Candidate Recycler}}
\label{sec:recycler}
\rev{The Recycler diagnoses and repairs the recoverable candidates selected by the Scheduler and returns them to the search pool.} For a weak-but-promising candidate, the goal is not to restart generation from random noise, but to reuse its clean latent as a structured prior and correct the diagnosed deficiency through prompt-guided re-denoising. Given a candidate $j\in\mathcal{W}_i$ at checkpoint $s_i$, \rev{the Recycler} takes the clean latent $x_0^j$ encoded from video $v^j$, the current prompt $p^j_{s_i}$, reward scores, and the stage-dependent focus $\rho_i$ assigned by \rev{the Scheduler} as inputs. We uniformly sample four keyframes $\mathcal{K}(v^j)$ from $v^j$ and use them as visual anchors for an MLLM $\mathcal{M}$ to generate an enhanced prompt $p^j_{s_{i+1}}$ with stage-specific instruction $I$ detailed in Appendix Sec. B:
\begin{equation}
    p^j_{s_{i+1}} = \mathcal{M} \bigl( p^j_{s_i}, \rho_i(v^j), \mathcal{K}(v^j), I \bigr).
\end{equation}
The output $p^j_{s_{i+1}}$ is a complete prompt that preserves the original semantics while emphasizing the current deficiency: motion dynamics in the high-noise regime and visual details in the low-noise regime. As established in Sec.~\ref{sec:method_overview}, this updated text will serve as the specific prompt condition for this candidate in the subsequent generation \rev{stage}. Since the rewriting is conditioned on both visual anchors and reward feedback, \rev{the Recycler operates on the existing candidate rather than acting as a prompt-only refinement method.}

\rev{The Recycler} then performs latent-space re-denoising under the enhanced prompt. It first perturbs the clean latent $x_0^j$ back to the active checkpoint $s_i$ using the same forward perturbation schedule as in Eq.~\eqref{eq:renoise}, yielding $\hat{x}_{s_i}^j$. Starting from $\hat{x}_{s_i}^j$, \rev{the Recycler} re-denoises the latent with the enhanced prompt using a manifold-aware SDE (MA-SDE) sampler. We denote the corresponding reverse process by $\mathcal{G}^{\mathrm{MA\mbox{-}SDE}}$, and the re-denoised latent is obtained by
\begin{equation}
    \hat{x}_0^j = \mathcal{G}^{\mathrm{MA\mbox{-}SDE}}_{s_i\rightarrow 0} 
    (\hat{x}_{s_i}^j,p^j_{s_{i+1}}),
\end{equation}
where the newly generated $\hat{x}_0^j$ replaces the original $x_0^j$, and the updated pair $(\hat{x}_0^j, p^j_{s_{i+1}})$ is merged into the survivor set $\mathcal{B}_i$ for the next stage. The use of MA-SDE is motivated by the need to balance exploration and preservation. During editing, the trajectory is already perturbed in two ways: the generated latent is re-noised, and the text condition is changed from $p^j_{s_i}$ to $p^j_{s_{i+1}}$. If the SDE solver injects excessive stochasticity on top of these perturbations, the sample may move away from the pretrained generator's data manifold \cite{karras2022elucidating,xu2023restart,chung2025cfg_pp}, which can lead to subject drift or background inconsistency.

Following \citet{zheng2026manifold}, for two consecutive solver noise levels $u_m>u_{m+1}$, the injected noise standard deviation is
\begin{equation}
    \Sigma_m^{1/2} = \eta_{\mathrm{sde}} 
    \sqrt{-(u_m-u_{m+1}) + \log\frac{1-u_{m+1}}{1-u_m}},
\end{equation}
where $\eta_{\mathrm{sde}}$ controls the amount of stochastic exploration. Unlike first-order SDE discretizations that estimate the noise variance with a linear approximation, MA-SDE integrates the diffusion coefficient exactly over each solver step, which reduces excess noise energy, thereby keeping the noise trajectory closer to the data manifold.

As illustrated in Fig.~\ref{fig:effect_of_recycler}, GEARS performs stage-aware candidate repair along the denoising trajectory, correcting motion and coarse structure at high-noise checkpoints while refining appearance and local details at low-noise checkpoints.
\begin{table}[h]
    \caption{\textbf{Generalization across Backbones}.}
    \label{tab:generalization}
    \begin{minipage}{\columnwidth}
    \centering
    \begin{tabular}{lcccc}
        \toprule
        & \multicolumn{2}{c}{\textbf{CogVideoX-2B}} & \multicolumn{2}{c}{\textbf{Wan2.1-VACE-1.3B}} \\
        \cmidrule(lr){2-3} \cmidrule(lr){4-5}
        \textbf{Method} & Quality & Semantic & Quality & Semantic \\
        \midrule
        Base & 0.8097 & 0.7639 & 0.8480 & 0.6573 \\
        $+$ Video-T1 & 0.8070 & 0.7679 & 0.8454 & 0.7415 \\
        $+$ DLBS & 0.8079 & 0.7716 & 0.8382 & 0.6985\\
        $+$ EvoSearch & 0.8006 & 0.7639 & 0.8703 & 0.6512 \\
        $+$ GEARS (\textbf{Ours}) & 0.8166 & 0.7973 & 0.8527 & 0.7616 \\
        \bottomrule
    \end{tabular}
    \end{minipage}
\end{table}
\begin{figure}[h]
    \centering
    \includegraphics[width=\columnwidth]{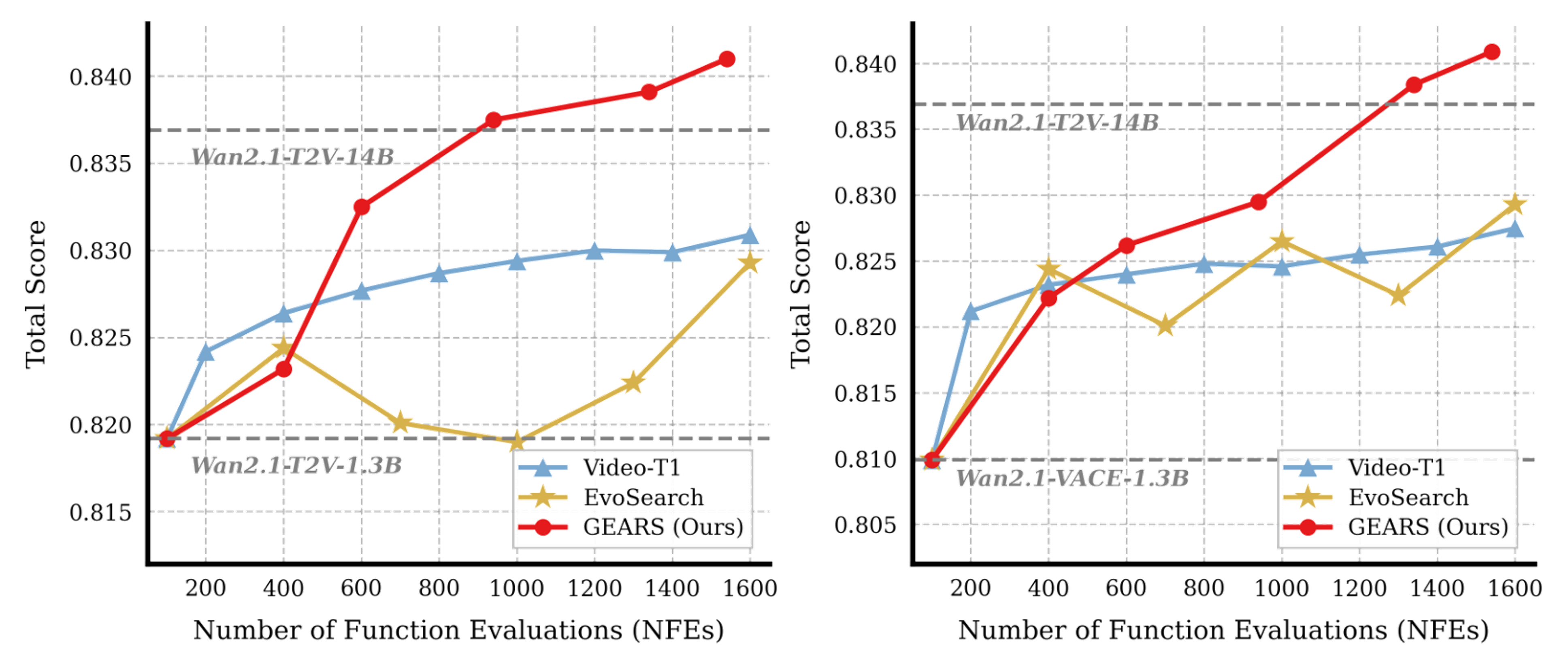}
    \caption{\textbf{Scaling behavior} on Wan2.1-T2V-1.3B and Wan2.1-VACE-1.3B.}
    \label{figure:scaling_behavior}
\end{figure}

\section{Experiments}\label{sec:experiments}
\subsection{Experiment Settings}

\paragraph{Implementation Details}
\rev{We evaluate GEARS on Wan2.1-T2V-1.3B/-14B \cite{wan2025wan}, Wan2.1-VACE-1.3B \cite{jiang2025vace}, and CogVideoX-2B \cite{yang2024cogvideox}. All Wan2.1 models generate 33-frame, approximately 2-second videos (81 frames for 5-second) at $832\times480$. CogVideoX-2B generates 49 frames at $720\times480$. We use dpmsolver$++$ for Wan2.1 and an SDE-augmented DDIM scheduler for CogVideoX.} \rev{VideoAlign \cite{liu2025improving} serves as the multi-dimensional reward model, and Qwen3.5-Plus \cite{bai2023qwen,qwen35} is the default MLLM in the Recycler. Our default configuration uses checkpoints $\{s_0=1,s_1=0.6,s_2=0.3\}$, transition index $\alpha=1$, $K_1=K_2=4$, and $q_{\mathrm{edit}}=0.5$, which recycles the bottom two valid candidates at each checkpoint. For NFE-matched comparisons \cite{liu2025video,he2025scaling}, we set the initial population $N_0$ to 10 for Video-T1 and DLBS \cite{oshima2025inference}, 8 for EvoSearch \cite{he2025scaling}, and 6 for GEARS. The same GEARS and candidate-budget configuration is used for Wan2.1-T2V-14B; detailed NFE accounting is provided in Appendix Sec. A.}

\paragraph{Evaluation Metrics and Protocol} \rev{We use VBench \cite{huang2024vbench} as the primary independent evaluation and report its total, quality, and semantic scores, together with background consistency, imaging quality, motion smoothness, and subject consistency. We additionally report VideoAlign visual- and motion-quality rewards to measure search efficiency in the reward space. Following \cite{oshima2025inference} and \cite{he2025scaling}, the evaluations use 110 prompts uniformly sampled across the VBench dimensions. We run five matched seeds per prompt for all locally evaluated methods, yielding 550 output videos per method.}

\subsection{Results}

\paragraph{Quantitative Results}
As shown in Table~\ref{tab:vbench_results}, GEARS achieves the highest Total Score among the compared TTS methods. On Wan2.1-T2V-1.3B, it improves the base model from $0.8192$ to $0.8375$, reaching a level comparable to or slightly above larger backbones such as the Wan2.1-T2V-14B ($0.8369$). The per-dimension results also suggest that different strategies introduce different trade-offs. For example, EvoSearch improves imaging quality but decreases motion smoothness and subject consistency, indicating that local noise perturbations may disrupt temporal or structural coherence. Video-T1 improves VideoAlign rewards but shows limited gains on several VBench dimensions, consistent with the reward-overfitting issue discussed in prior work \cite{liu2025video}. In contrast, GEARS achieves stronger overall performance by recycling low-scoring but recoverable candidates through diagnosis-conditioned editing.

\paragraph{Generalization across Different Backbones}
We further evaluate GEARS on different model families, including CogVideoX-2B \cite{yang2024cogvideox} and Wan2.1-VACE-1.3B \cite{jiang2025vace}. As shown in Table~\ref{tab:generalization}, GEARS consistently improves semantic scores and maintains stable quality gains across both backbones, demonstrating its applicability across different backbones.

\paragraph{Scaling Behavior at Test-Time}
Fig.~\ref{figure:scaling_behavior} compares the scaling behavior of GEARS and baseline methods on Wan2.1-T2V-1.3B and Wan2.1-VACE-1.3B. At small compute budgets (e.g., NFEs=400, corresponding to four initial videos), GEARS and EvoSearch underperform Video-T1, suggesting that directly allocating compute to sample more initial candidates is more effective when the candidate pool is limited. As the compute budget increases, however, GEARS exhibits more efficient performance gains, indicating that recycling and editing partially successful candidates becomes increasingly effective once sufficient candidate diversity is available.

\begin{table}[t]
\revcolor
\caption{\textbf{Evaluation on 5-Second Videos and a Larger Backbone.}}
\label{tab:longer_larger}
\centering
\begin{minipage}[t]{0.48\columnwidth}
\centering
(a) 5-Second Videos
\begin{tabular*}{\linewidth}{@{\extracolsep{\fill}}lc@{}}
    \toprule
    \textbf{Method} & \textbf{Total} \\
    \midrule
    Wan2.1-T2V-1.3B & 0.8207 \\
    $+$ Video-T1 & 0.8272 \\
    $+$ EvoSearch & 0.8261 \\
    $+$ GEARS (\textbf{Ours}) & \textbf{0.8371} \\
    \bottomrule
\end{tabular*}
\end{minipage}
\hfill
\begin{minipage}[t]{0.48\columnwidth}
\centering
(b) Larger Backbone
\begin{tabular*}{\linewidth}{@{\extracolsep{\fill}}lc@{}}
    \toprule
    \textbf{Method} & \textbf{Total} \\
    \midrule
    Wan2.1-T2V-14B & 0.8369 \\
    $+$ Video-T1 & 0.8410 \\
    $+$ EvoSearch & 0.8388 \\
    $+$ GEARS (\textbf{Ours}) & \textbf{0.8425} \\
    \bottomrule
\end{tabular*}
\end{minipage}
\end{table}

\begin{table}[t]
\revcolor
\caption{\textbf{Prompt-level stability and blinded human evaluation of Wan2.1-T2V-1.3B $+$ GEARS.} Prompt-level results average five independent seeds for each of the 110 prompts; human results report Win/Tie/Loss.}
\label{tab:stability_human}
\centering
\small
\setlength{\tabcolsep}{3pt}
\begin{tabular}{lccc}
    \toprule
    \textbf{Compared Method} & $\Delta$ \textbf{Score} & \makecell{\textbf{Prompt}\\\textbf{Wins}} & \makecell{\textbf{Human}\\\textbf{W/T/L}} \\
    \midrule
    Wan2.1-T2V-1.3B & $+$0.018 & 97/110 & 75/11/14 \\
    $+$ Video-T1 & $+$0.007 & 79/110 & 68/14/18 \\
    $+$ EvoSearch & $+$0.019 & 85/110 & 85/3/12 \\
    Wan2.1-T2V-14B & $+$0.004 & 71/110 & 68/3/29 \\
    \bottomrule
\end{tabular}
\end{table}

\begin{table}[t]
\revcolor
\caption{\textbf{End-to-end inference cost.} PFLOPs are measured with a local Qwen3.5-9B. GEARS wall-clock time includes MLLM API latency.}
\label{tab:compute_cost}
\centering
\small
\setlength{\tabcolsep}{3pt}
\begin{tabular}{lcc}
    \toprule
    \textbf{Method} & \textbf{PFLOPs} & \makecell{\textbf{Time (s)}\\\textbf{33 frames}} \\
    \midrule
    Wan2.1-T2V-14B & 50.53 & $200.37\pm3.04$ \\
    Wan2.1-T2V-1.3B $+$ GEARS & 52.77 & $189.66\pm4.91$ \\
    \bottomrule
\end{tabular}
\end{table}

\begin{figure*}[!tb]
    \centering
    \includegraphics[width=\linewidth]{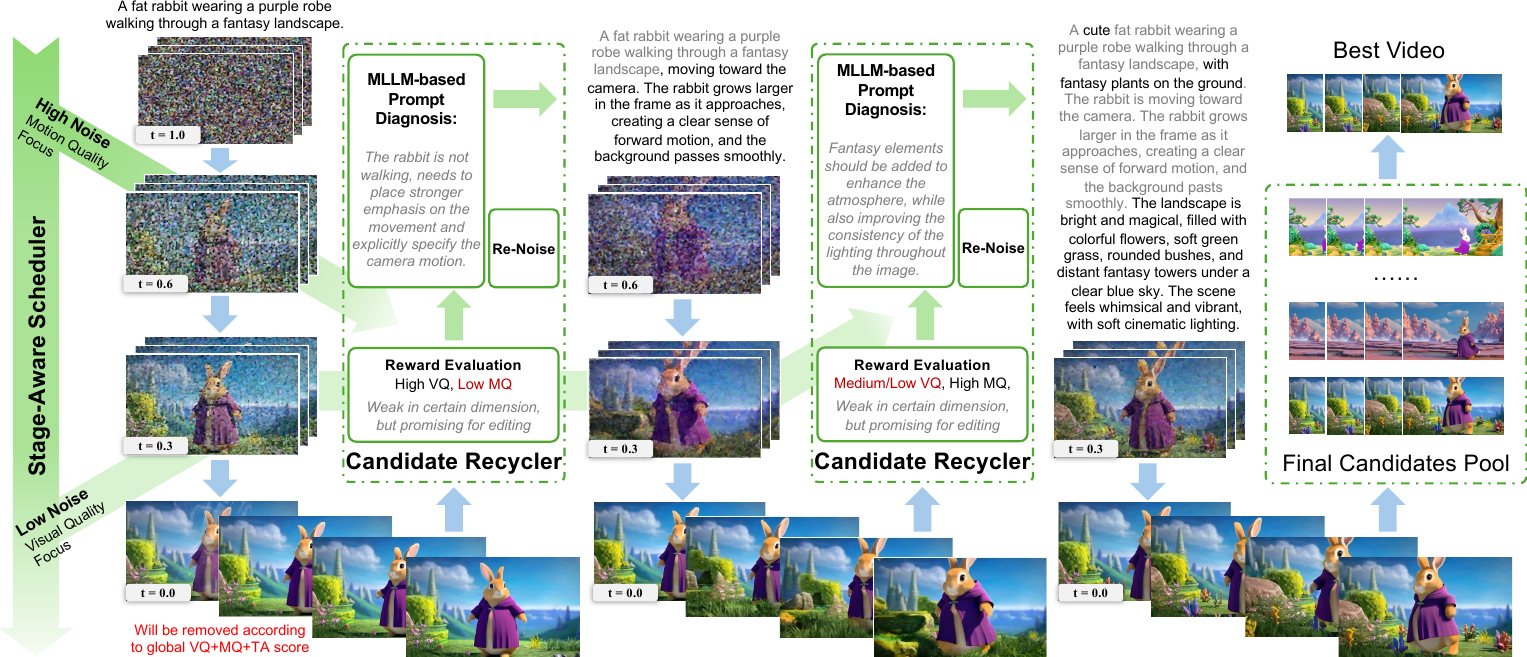}
    \caption{\textbf{Discarded-to-Improved Trajectory in GEARS.} A low-ranked but semantically valid candidate, which would be discarded by conventional generate-and-select TTS, is recycled through \rev{the Scheduler and Recycler}. GEARS first repairs its motion deficiency at a high-noise checkpoint and later enhances visual details at a low-noise checkpoint. The repaired candidate is then returned to the final candidate pool and selected as the best video.}\label{fig:overall_process_rabbit}
\end{figure*}

\begin{figure*}[!tb]
    \centering
    \includegraphics[width=\linewidth]{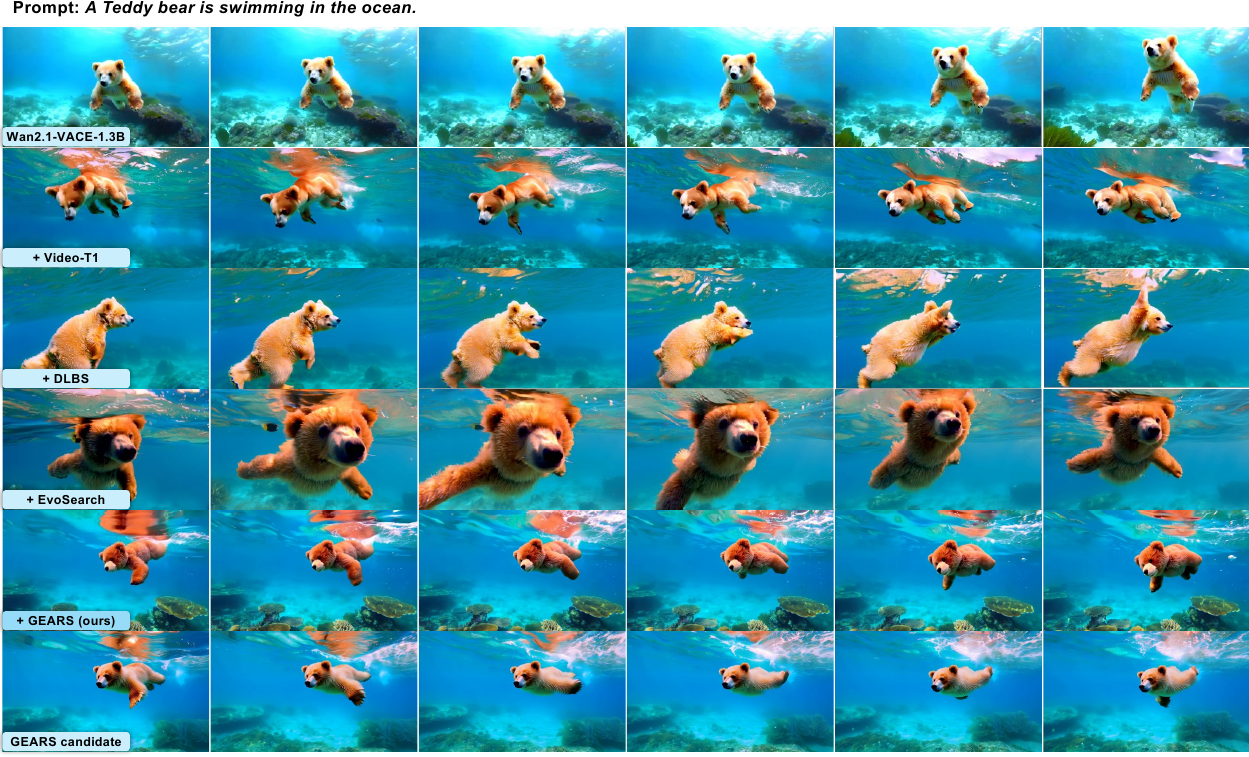}
    \caption{Visualized examples on Wan2.1-VACE-1.3B.}\label{visualization_on_vace}
\end{figure*}

\rev{\paragraph{Longer-Video Generation} We use the approximately 2-second setting above to follow recent video TTS methods \cite{he2025scaling,wu2026imagerysearch} and enable broad comparisons under our compute budget; GEARS itself is not restricted to this duration. On 5-second Wan2.1-T2V-1.3B videos, GEARS reaches a VBench Total of 0.8371 and outperforms the base model, Video-T1, and EvoSearch under comparable budgets (Table~\ref{tab:longer_larger}(a)).}

\rev{\paragraph{Scaling to a Larger Backbone} We further apply GEARS to Wan2.1-T2V-14B using the same evaluation protocol and candidate-budget configuration as the main experiment. As shown in Table~\ref{tab:longer_larger}(b), GEARS improves the 14B base model from 0.8369 to 0.8425, outperforming Video-T1 (0.8410) and EvoSearch (0.8388). This result shows that candidate recycling remains effective on a strong large-scale backbone, beyond helping a smaller model close the quality gap.}

\rev{\paragraph{Prompt- and Seed-Level Stability} After averaging the five seeds for each prompt, Wan2.1-T2V-1.3B with GEARS improves the normalized prompt-level score over the 1.3B base model, Video-T1, and EvoSearch by 0.018, 0.007, and 0.019, respectively, and wins on 97, 79, and 85 of the 110 prompts (Table~\ref{tab:stability_human}). It also outperforms the Wan2.1-T2V-14B base model by 0.004 and wins on 71 of the 110 prompts. These paired results indicate that the gains are not driven by a particular prompt subset or initial noise pool.}

\rev{\paragraph{Human Evaluation and Reward Robustness} We conduct a blinded pairwise study in which participants view the original prompt and two anonymized videos in randomized order and select ``prefer A,'' ``prefer B,'' or ``tie.'' We collect 400 judgments from 20 participants over 20 prompts, with one comparison per prompt for each participant. As shown in the last column of Table~\ref{tab:stability_human}, GEARS obtains 296/31/73 wins/ties/losses overall. GEARS uses VideoAlign-derived rewards for routing and final ranking, whereas VBench and human preferences are used only for evaluation. The consistent gains under these independent evaluations suggest that the improvements are not solely an artifact of optimizing the routing reward.}

\rev{\paragraph{End-to-End Compute} Table~\ref{tab:compute_cost} complements NFE accounting with end-to-end FLOPs and wall-clock time. Under the setting of Table~\ref{tab:vbench_results}, GEARS uses approximately 52.77 PFLOPs: 51.92 for the Wan2.1-1.3B generator, 0.70 for VideoAlign, and approximately 0.15 for a local Qwen3.5-9B. This is about 1.04$\times$ the 50.53 PFLOPs of Wan2.1-14B. On one NVIDIA H100, generating 33 frames takes $189.66\pm4.91$ seconds for GEARS and $200.37\pm3.04$ seconds for Wan2.1-14B; the GEARS measurement uses the MLLM API and includes its latency.}

\begin{figure*}[!htbp]
    \centering
    \includegraphics[width=0.95\linewidth]{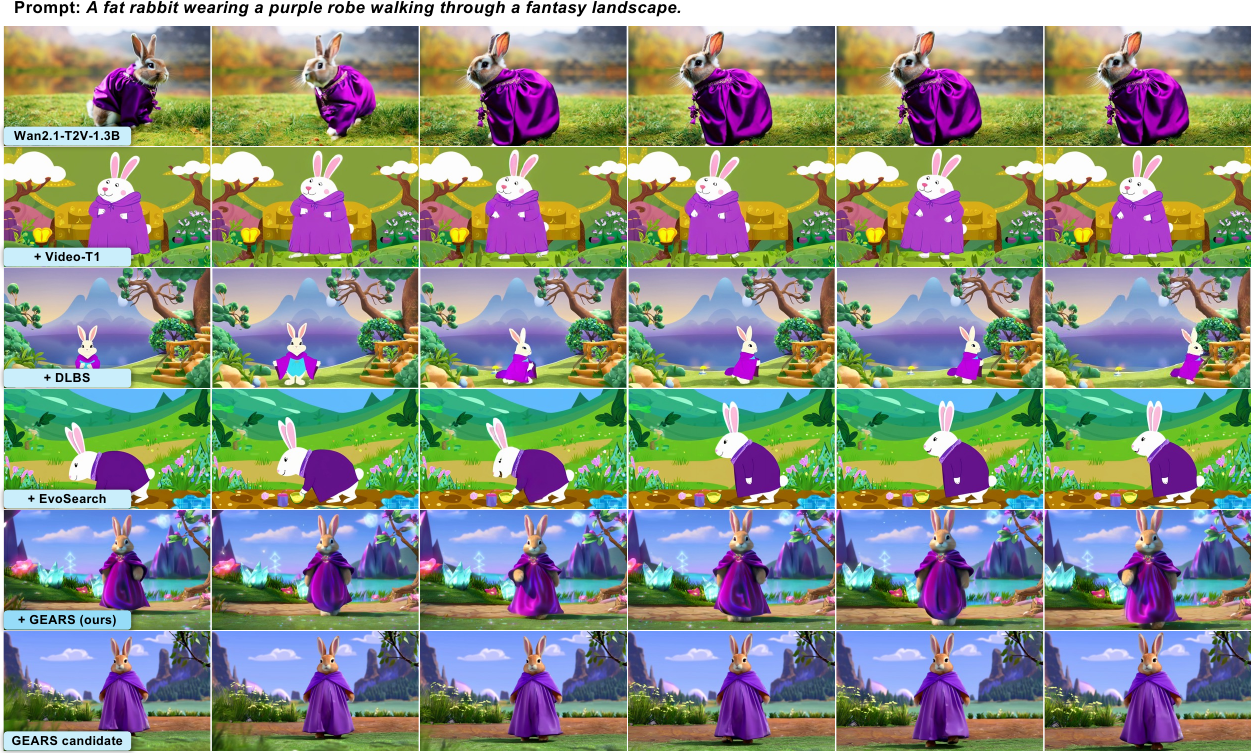}
    \includegraphics[width=0.95\linewidth]{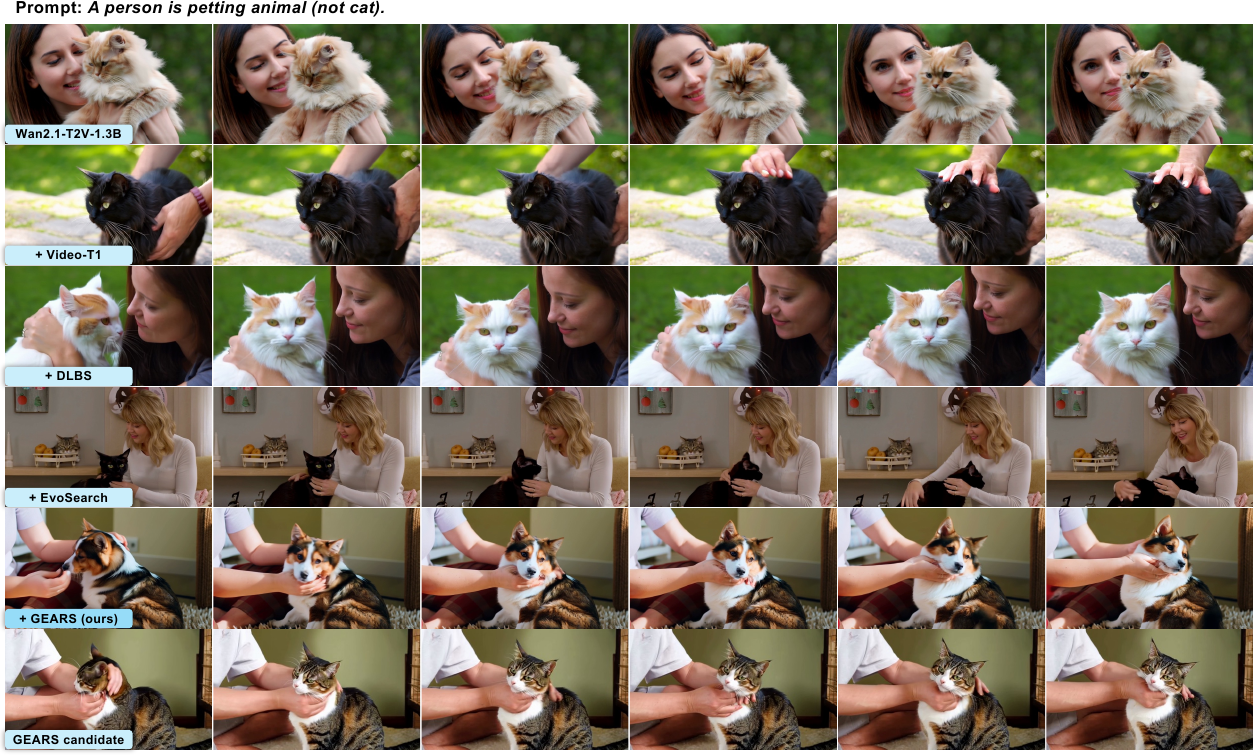}
    \caption{Visualized examples on Wan2.1-T2V-1.3B.}\label{visualization_on_want2v}
\end{figure*}

\paragraph{Qualitative Explanation and Comparison}
Qualitative results illustrate how GEARS converts low-ranked but recoverable candidates into competitive outputs. In Fig.~\ref{fig:effect_of_recycler}, the \rev{Stage} 1 rabbit candidate preserves its subject identity and visual quality but under-realizes the requested walking motion. \rev{The Scheduler} therefore routes it to high-noise motion repair, where \rev{the Recycler} produces coherent whole-body displacement; the background-compensated optical-flow visualization further corroborates the stronger subject motion. Conversely, the \rev{Stage} 2 beach candidate already exhibits plausible layout and wave dynamics but lacks the requested cyberpunk appearance. \rev{The Scheduler} routes it to low-noise visual repair, allowing \rev{the Recycler} to introduce neon structures while retaining the coastal composition. Fig.~\ref{fig:overall_process_rabbit} further traces the rabbit candidate through the complete recycling loop. Rather than being removed because of its low MQ and consequently low global rank, the candidate is repaired, returned to the candidate pool, and ultimately selected as the best video. Fig.~\ref{fig:teaser}, Fig.~\ref{visualization_on_vace}, and Fig.~\ref{visualization_on_want2v} provide additional qualitative comparisons with existing TTS methods.

\begin{table}[t]
\caption{\textbf{Ablation of \rev{Stage-Aware Scheduler} Settings.} $\alpha$ is the transition index that partitions the checkpoints into high- and low-noise stages.}
\label{tab:stageadapter_ablation}
\centering
\small
\setlength{\tabcolsep}{3pt}
\begin{tabular}{lcc}
    \toprule
    \textbf{Checkpoint Setting} & \textbf{Focus} & \textbf{Total Score}\\
    \midrule
    $\{s_0=1\}$ & No adapt & 0.8192\\
    $\{s_0=1, s_1=0.6\}$ & MQ only & 0.8221\\
    $\{s_0=1, s_1=0.3\}$ & VQ only & 0.8239\\
    $\{s_0=1, s_1=0.6, s_2=0.3\}$ & $\alpha=1$ & 0.8375\\
    $\{s_0=1, s_1=0.6, s_2=0.45, s_3=0.3\}$ & $\alpha=1$ & 0.8304\\
    $\{s_0=1, s_1=0.6, s_2=0.45, s_3=0.3\}$ & $\alpha=2$ & 0.8392\\
    $\{s_0=1, s_1=0.75, s_2=0.6, s_3=0.3\}$ & $\alpha=1$ & 0.8379\\
    $\{s_0=1, s_1=0.75, s_2=0.6, s_3=0.3\}$ & $\alpha=2$ & 0.8410\\
    \bottomrule
\end{tabular}
\end{table}

\subsection{Ablation Studies}
\paragraph{Ablation of \rev{Stage-Aware Scheduler} Settings.}
Table~\ref{tab:stageadapter_ablation} evaluates \rev{stage-aware} checkpoint scheduling. Without adaptive recycling, the base model obtains $0.8192$. Adding one editing checkpoint yields limited gains, reaching $0.8221$ for motion-focused editing at $s_1=0.6$ and $0.8239$ for visual-focused editing at $s_1=0.3$. Furthermore, combining a high-noise motion checkpoint with a low-noise visual checkpoint improves the score to $0.8375$, showing that different failure modes benefit from correction at their corresponding denoising \rev{stages}. Increasing the number of checkpoints can further improve performance when their \rev{stage assignments} follow the denoising hierarchy, but also incurs additional inference cost: for $\{0.6,0.45,0.3\}$, treating the first two checkpoints as motion-oriented improves the score from $0.8304$ to $0.8392$, and the setting $\{0.75,0.6,0.3\}$ with $\alpha=2$ achieves the best score of $0.8410$. These results support \rev{the Scheduler} from two aspects: \rev{stage-aware} scheduling is both necessary and effective, and checkpoint granularity offers a controllable trade-off between test-time computation and generation quality.

\rev{\paragraph{Sensitivity to the Editing Ratio} We vary $q_{\mathrm{edit}}$ over $0.25/0.50/0.75$, which requires $810/900/990$ NFEs and yields VBench Total scores of $0.8359/0.8375/0.8384$, respectively. All settings remain within the 1000-NFE baseline budget. The stable scores indicate that GEARS is not sensitive to the editing ratio, while allocating more compute to recycling provides a modest improvement.}

\paragraph{Ablation of \rev{Candidate Recycler} Components.}
Table~\ref{tab:component_ablation} ablates the two main components of \rev{the Recycler} under \rev{the Scheduler} with checkpoints $\{s_0=1, s_1=0.6, s_2=0.3\}$ and $\alpha=1$. Starting from the base score of $0.8192$, adding MLLM diagnosis with the default SDE sampler improves the score to $0.8281$ with Qwen3.5-9B and $0.8325$ with Qwen3.5-Plus, showing that reward- and frame-conditioned diagnosis helps turn weak-but-promising candidates into useful editing targets. Replacing the standard SDE sampler with MA-SDE further improves the score to $0.8357$ and $0.8375$, which improves subject consistency and motion smoothness, suggesting that more stable re-denoising is essential after the editing condition changes to reduce subject drift and temporal jitter. The combination of MLLM diagnosis and MA-SDE achieves the best performance, demonstrating that both components are necessary for effective \rev{recycling}.

\begin{table}[t]
\caption{\textbf{Ablation of \rev{Candidate Recycler} Components.}}
\label{tab:component_ablation}
\centering
\small
\setlength{\tabcolsep}{3pt}
\begin{tabular}{cccccc}
\toprule
\multicolumn{2}{c}{\textbf{MLLM Diagnosis}} & \multirow{2}{*}{\textbf{MA-SDE}} & \multicolumn{3}{c}{\textbf{VBench}} \\
\cmidrule(lr){1-2} \cmidrule(lr){4-6}
Enabled & Model & & Total & Sub. Con. & Mot. Smo. \\
\midrule
$\times$ & - & $\times$ & 0.8192 & 0.9687 & 0.9674 \\
\checkmark & Qwen3.5-9B & $\times$ & 0.8281 & 0.9782 & 0.9689 \\
\checkmark & Qwen3.5-Plus & $\times$ & 0.8325 & 0.9807 & 0.9692 \\
\checkmark & Qwen3.5-9B & \checkmark & 0.8357 & 0.9812 & 0.9715 \\
\checkmark & Qwen3.5-Plus & \checkmark & 0.8375 & 0.9815 & 0.9727 \\
\bottomrule
\end{tabular}
\end{table}

\section{Conclusion, Limitations and Future Work}\label{sec:conclusion}
We introduced \rev{GEARS (Guided Editing for Adaptive Recycling Search)}, \rev{a test-time scaling framework for video generation that shifts inference-time search from passive noise-space exploration to active candidate recycling.} Instead of treating low-scoring samples as failed trials, GEARS identifies semantically plausible candidates with recoverable deficiencies, repairs them at denoising \rev{stages} where the corresponding attributes are most controllable, and \rev{returns them to} the search process. \rev{By converting otherwise discarded candidates into structured refinement paths, this generation-to-editing transition enables more effective use of inference-time compute.} Experiments show that GEARS consistently improves generation quality compared to baseline methods. GEARS is most effective when low-scoring candidates remain semantically plausible and preserve recoverable structure. Severe text misalignment or multiple entangled failures can still be difficult to repair reliably. In addition, the current checkpoint schedule and transition boundary are empirically chosen based on the denoising hierarchy, which may not be optimal for every backbone, prompt, or budget. Future work includes developing more efficient latent-space evaluation and diagnosis mechanisms, learning adaptive checkpoint schedules and budget allocation policies, and extending candidate recycling to longer videos and broader generation-editing scenarios.

\begin{acks}
This work was supported by National Key Research and Development Program of China (2025YFA1805700); National Natural Science Foundation of China (82371112); Science Foundation of Peking University Cancer Hospital (JC202505).
\end{acks}

\bibliographystyle{ACM-Reference-Format}
\bibliography{sample-bibliography}